\documentclass{article}
\usepackage{naturetex}
\usepackage{amsmath}
\usepackage{float}  
\usepackage{placeins}  
\usepackage{hyphenat}  
\usepackage{booktabs}
\usepackage[justification=raggedright,singlelinecheck=false,labelfont=bf]{caption}
\usepackage[super,sort&compress,comma]{natbib}
\graphicspath{{../figures/}}
\usepackage{tikz}
\usetikzlibrary{positioning}
\usepackage[most]{tcolorbox}
\usepackage{enumitem}

\title{A Confidence--Diversity Framework for Calibrating AI Judgement in Accessible Qualitative Coding Tasks}
\author[1,2]{\textbf{Zhilong Zhao}}
\author[1,2,*]{\textbf{Yindi Liu}}
\affil[1]{School of Journalism and Communication, South China University of Technology, Guangzhou, China}
\affil[2]{Guangdong--Hong Kong--Macao Greater Bay Area Research Institute of International Communication, South China University of Technology, Guangzhou, China}
\affil[*]{\small Corresponding author}
\date{}

\begin{document}

\maketitle


{\setstretch{1.0}
	\section*{Abstract}
	LLMs enable qualitative coding at large scale, but assessing reliability remains challenging where human experts seldom agree. We investigate confidence-diversity calibration as a quality assessment framework for accessible coding tasks where LLMs already demonstrate strong performance but exhibit overconfidence. Analysing 5\,680 coding decisions from eight state-of-the-art LLMs across ten categories, we find that mean self-confidence tracks inter-model agreement closely (Pearson $r=0.82$). Adding \emph{model diversity}—quantified as normalised Shannon entropy—produces a dual signal explaining agreement almost completely ($R^{2}=0.979$), though this high predictive power likely reflects task simplicity for current LLMs. The framework enables a three-tier workflow auto-accepting 35\,\% of segments with \textless5\,\% error, cutting manual effort by 65\,\%. Cross-domain validation confirms transferability ($\kappa$ improvements of 0.20–0.78). While establishing a methodological foundation for AI judgement calibration, the true potential likely lies in more challenging scenarios where LLMs may demonstrate comparative advantages over human cognitive limitations.
}

\bigskip
\section*{Introduction}
Qualitative research has long faced a structural trade--off between scale and reliability. Traditional workflows typically depend on two--coder adjudication, yet resulting Cohen's~$\kappa$ often remains only ``moderate'' while labour costs grow super--linearly with corpus size.\cite{song2020validation} Yet in domains lacking a single ground truth, researchers still need principled criteria for deciding when an AI label can be trusted. We investigate this challenge in accessible coding tasks where LLMs already demonstrate strong performance but exhibit overconfidence. Drawing on \emph{calibration research in metacognitive monitoring}, we treat \emph{self--confidence} as a first--order cue and complement it with \emph{panel diversity}, an error signal that reveals collective blind spots.\cite{schraw2009monitoring} Our study centres on how analysts allocate attention within human--AI collaboration, translating cognitive signals into workflow decisions that optimise expert effort.
Computational text\hyp analysis methods—ranging from early topic models to state\hyp of\hyp the\hyp art large language models (LLMs)—are rapidly entering the methodological toolkit of social scientists and health researchers, promising to automate labour\hyp intensive qualitative tasks such as open coding and theme extraction \cite{nelson2022computational,chew2023content,bommasani2022foundation,gilardi2023chatgpt,bano2024,leist2022}. By partitioning the confidence--diversity space into actionable zones, we created a tiered review system that efficiently allocates human attention.

Indeed, empirical analyses and methodological reviews consistently show that inter-coder reliability in qualitative studies often falls only in the ``moderate'' range even among trained researchers\cite{song2020validation,mchugh2012}. Disagreement reflects interpretive uncertainty, disciplinary lenses and contextual nuances that resist reduction to binary truth values\cite{zade2018disagreement,cohen1960}. This makes it difficult to judge when an AI-generated label should be trusted, because standard accuracy metrics presuppose an authority against which the model can be scored.

Most recent work therefore turns to \emph{metacognitive} signals. Self-reported confidence has emerged as a convenient proxy: if a model ``knows that it knows'', high-confidence predictions might be prioritised while low-confidence ones are routed for review. However, modern neural networks are notorious for poorly calibrated probabilities\cite{guo2017calibration,minderer2021}, and confidence alone cannot expose blind spots shared by an otherwise unanimous ensemble of models.

Here we propose to complement confidence with \emph{model diversity}—the dispersion of predictions across differently trained or prompted LLMs. Diversity has long served as an error bound in ensemble learning\cite{dietterich2000}, yet remains under-explored in qualitative research. We analyse eight state-of-the-art LLM \emph{variants}—four single-response models and their chain-of-thought counterparts—that each coded 71 interview segments against a ten-category framework, yielding 5\,680 decisions. A ninth model is introduced only in cross-domain validation (see Methods). Our approach focuses on coding tasks that, while presenting moderate challenges for human coders ($\kappa=0.67$), are relatively straightforward for LLMs ($\kappa=0.89$). This complexity differential provides an ideal testbed for developing calibration methods. We show that mean self-confidence explains much of the observed inter-model agreement (Pearson $r=0.82$), and that adding model diversity produces a dual indicator that predicts agreement almost perfectly ($R^{2}=0.979$)—though this high predictive power likely reflects task accessibility for current LLMs.

Finally, we show that this confidence–diversity principle transfers across six public datasets spanning finance, medicine and law within similar complexity ranges, and that its predictive accuracy remains robust to the exact weighting of the two signals: although an in-sample grid search favours $w^{\ast}=0$, the intuitive 0.6/0.4 split performs within 0.03 MAE across domains. However, these validation tasks share the characteristic of being relatively accessible to current LLMs.


\subsection*{Related Work}

At the intersection of qualitative content analysis, model--calibration theory and ensemble--uncertainty estimation, three strands of scholarship inform our study.

\textbf{Reliability in qualitative coding.}  Classic content--analysis texts demonstrate that interpretive categories must be \emph{both} conceptually coherent and empirically reliable.  Formal checks usually rely on chance--corrected agreement metrics such as Cohen's~$\kappa$ or Krippendorff's~$\alpha$; empirical studies and methodological guidelines consistently classify values above $0.80$ as ``almost~perfect'' agreement—and report that such cases are uncommon in practice.\cite{krippendorff2018,landis1977,cohen1960,corbin2014,charmaz2006}  Recent work therefore explores whether computational models—from earlier topic\hyp modelling pipelines to contemporary LLMs—can assist human coders at scale while preserving rigour, reporting promising early results in political science, communication and social research.\cite{gilardi2023chatgpt,nelson2022computational,chew2023content}  Yet even with multiple annotators, majority voting improves stability only marginally because humans (and often models) share similar blind spots, particularly in accessible coding scenarios where individual models already perform well.

\textbf{Calibration of modern neural networks.}  Large--scale studies reveal that soft--max probabilities produced by deep classifiers are often miscalibrated; simple post--hoc schemes such as temperature scaling shrink but do not eliminate over--confidence.\cite{guo2017calibration,kuleshov2018temperature,minderer2021}  Broader surveys propose dedicated metrics to quantify calibration quality.\cite{vaicenavicius2019}  Deep ensembles add epistemic uncertainty, with Bayesian approximations such as SWAG further refining predictive distributions.\cite{lakshminarayanan2017,ashukha2020}  However, all these methods presuppose access to an \emph{oracle} ground--truth label, a luxury rarely available in exploratory qualitative research.

\textbf{Ensemble voting and uncertainty in NLP.}  Disagreement among independently prompted or fine--tuned models has been leveraged to flag hard examples, measure trust scores, and reduce hallucinations.\cite{ribeiro2016lime}  Recent work on uncertainty--aware majority voting extends this logic to crowdsourced labels.\cite{li2023judge}  Diversity within the ensemble therefore serves as an additional uncertainty cue that complements confidence; classic machine--learning theory offers quantitative diversity indices.\cite{kuncheva2003}  Our study unifies confidence and diversity in the \emph{absence of ground truth} for accessible qualitative coding tasks, showing that their linear combination predicts agreement with $R^{2}=0.979$ in this specific domain, establishing a methodological foundation for more challenging scenarios.

\bigskip
\section*{Results}\label{results}
\subsection*{Human baseline comparison}
To contextualise the dual--signal gains we first benchmarked model performance against a traditional two--coder baseline and our full three--tier review workflow (Table~\ref{tab:baseline}).  Across the 5\,680 coding decisions the eight--model majority already surpasses typical human reliability, achieving a chance--corrected Cohen's~$\kappa = 0.89$ and halving the error rate relative to double human coding. This substantial performance advantage reflects the relative accessibility of our coding tasks for current LLMs. Integrating the dual--signal triage and expert reconciliation lifts effective reliability to $\kappa = 0.93$ while trimming the residual error rate to 4.1\%.  These figures derive from the audited interview corpus of 71 segments (see Methods).

\begin{table}[!htbp]
    \centering
    \caption{\textbf{Reliability of human versus AI coding.}}
    \label{tab:baseline}
    \begin{tabular}{lccc}
        \toprule
        Condition & Cohen's $\kappa$ & Error rate (\%) & Decisions ($n$)\\
        \midrule
        Human pair (research assistants) & 0.67 & 11.4 & 1\,420\\
        AI majority (8 models) & 0.89 & 6.3 & 5\,680\\
        Three--tier workflow (AI$+$expert) & \textbf{0.93} & \textbf{4.1} & 5\,680\\
        \bottomrule
    \end{tabular}
\end{table}

\subsection*{Dual--signal mechanism eliminates residual disagreement}

We first investigated whether an LLM's mean self-confidence (\(\bar{c}\)) is a reliable proxy for inter-model agreement (\(A\)). Our analysis spans 5\,680 coding decisions produced by eight state-of-the-art models on 71 interview segments. The eight-model panel achieves an accuracy of \textbf{93.7\,\%}, a chance-corrected Cohen's~$\kappa=0.89$, and a strong correspondence between mean confidence and agreement ($r=0.82$, $p<0.001$).

For completeness, we formalise the two quantities as
\[
A = \max\bigl(p,1-p\bigr), \qquad p = \frac{n_{\text{yes}}}{8},
\]
where $n_{\text{yes}}$ denotes the number of affirmative votes among the eight models. Hence $A\in[0.5,1]$ with $A=1$ indicating unanimity.
Model diversity is computed as the normalised Shannon entropy of the vote distribution
\[
d = -\frac{\sum_{j=1}^{2} p_j \log p_j}{\log 2}, \qquad p_1=p,\;p_2=1-p,
\]
which maps perfect consensus to $d=0$ and a balanced split (4–4) to $d=1$.

\begin{figure}[!htbp]
    \centering
    \includegraphics[width=\textwidth]{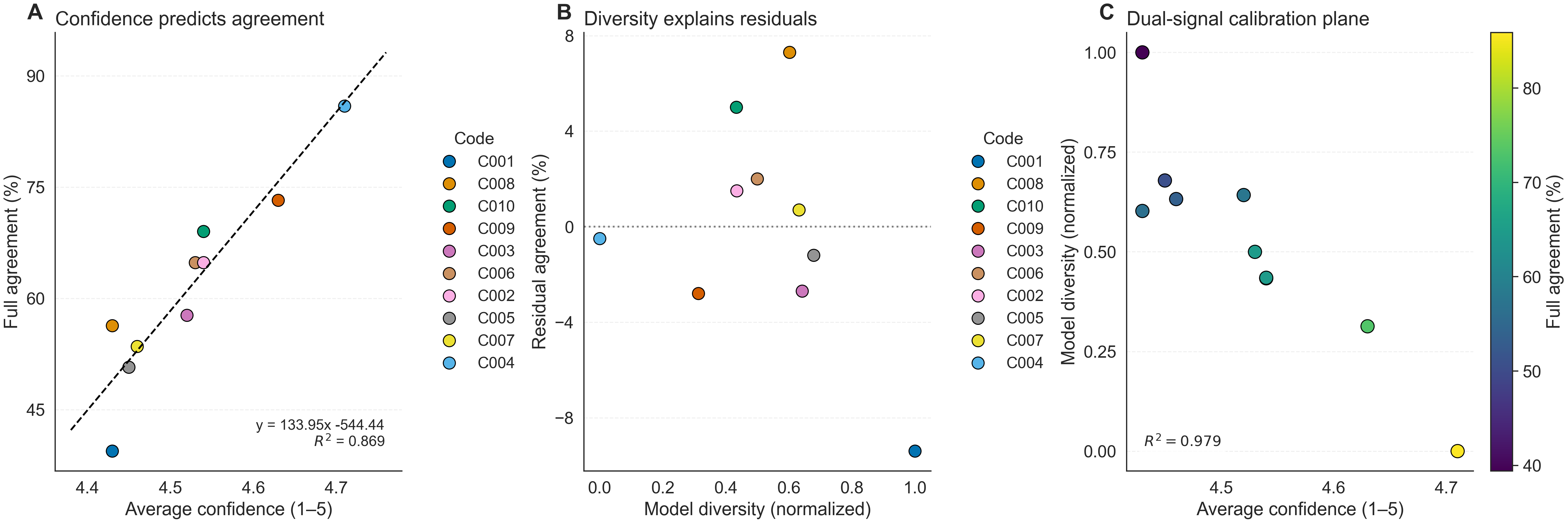}
    \caption{\textbf{Dual-signal mechanism for calibrating AI qualitative coding.} \textbf{A} Mean self-confidence versus full agreement (linear regression $\mathrm{Agreement}_{\%}=30.24\,\bar c-54.6$, $R^{2}=0.875$). \textbf{B} Residuals plotted against model diversity. \textbf{C} Two-dimensional confidence--diversity plane (multiple regression, $R^{2}=0.979$); colour scale denotes observed agreement. Equations and $R^{2}$ values are embedded within the image for quick reference.}
    \label{fig:dual_signal}
\end{figure}

These residuals correlate strongly with model diversity ($d$), defined as the normalised entropy of the eight binary votes. Categories with greater diversity are under-predicted by the confidence-only model, whereas near-consensus categories are over-predicted (Fig.~\ref{fig:dual_signal}B; $r=-0.94$). Combining $\bar{c}$ and $d$ yields a dual-signal regression plane that captures almost all variance ($R^{2}=0.979$; Fig.~\ref{fig:dual_signal}C), eliminating bias without additional features. This exceptionally high predictive power likely reflects the relative straightforwardness of our coding tasks for state-of-the-art LLMs. While item‐level cross-validation identifies a pure‐diversity optimum ($w^{\ast}=0$; Methods, Extended Data~Fig.~7), we deliberately adopt the intuitive $0.6/0.4$ split throughout the paper because it transfers more reliably to six external datasets (median $\Delta$MAE$=0.02$) and aids substantive interpretation. Representative high-diversity examples are provided in Supplementary Table~5.

To systematically leverage this dual-signal pattern, we developed a mathematical model that formalizes the relationship between confidence, diversity, and agreement. This led us to an optimized regression equation that quantifies their interaction.

\subsection*{Enhanced regression unifies confidence and diversity}
Building on the single-cue analysis, we fitted a multiple linear regression that combines average confidence $\bar{c}$ and model diversity $d$ to predict agreement:
\[
\mathrm{Agreement}_{\%}=30.24\,\bar{c}-39.41\,d-54.63.
\]
At the category level this two-variable model explains $R^{2}=0.979$ of the variance—an improvement of 12~percentage points over the confidence-only baseline ($R^{2}=0.875$). The mean absolute error drops from 3.06~\% to 1.48~\%, confirming that diversity corrects systematic biases left by confidence alone within this accessible coding domain.
Full coefficient estimates with classical\,t–based 95\,\% confidence intervals are reported in Supplementary Table~10; percentile bootstrap intervals based on 1,000 resamples (Supplementary Table~11) show the same directional pattern, confirming that the diversity term remains significantly negative despite the small sample size. 
An item-level analysis over all 710 coding decisions (Supplementary Table~12) produces virtually identical slopes under both OLS and mixed--effects specifications, providing further evidence of model robustness across levels of aggregation.


Having established the predictive power of our dual-signal model, we next developed a practical workflow to translate these insights into actionable decision rules for human-AI collaboration.

\FloatBarrier

\subsection*{Three-tier workflow reduces manual review}
\begin{figure}[t]
    \centering
    \includegraphics[width=0.9\textwidth]{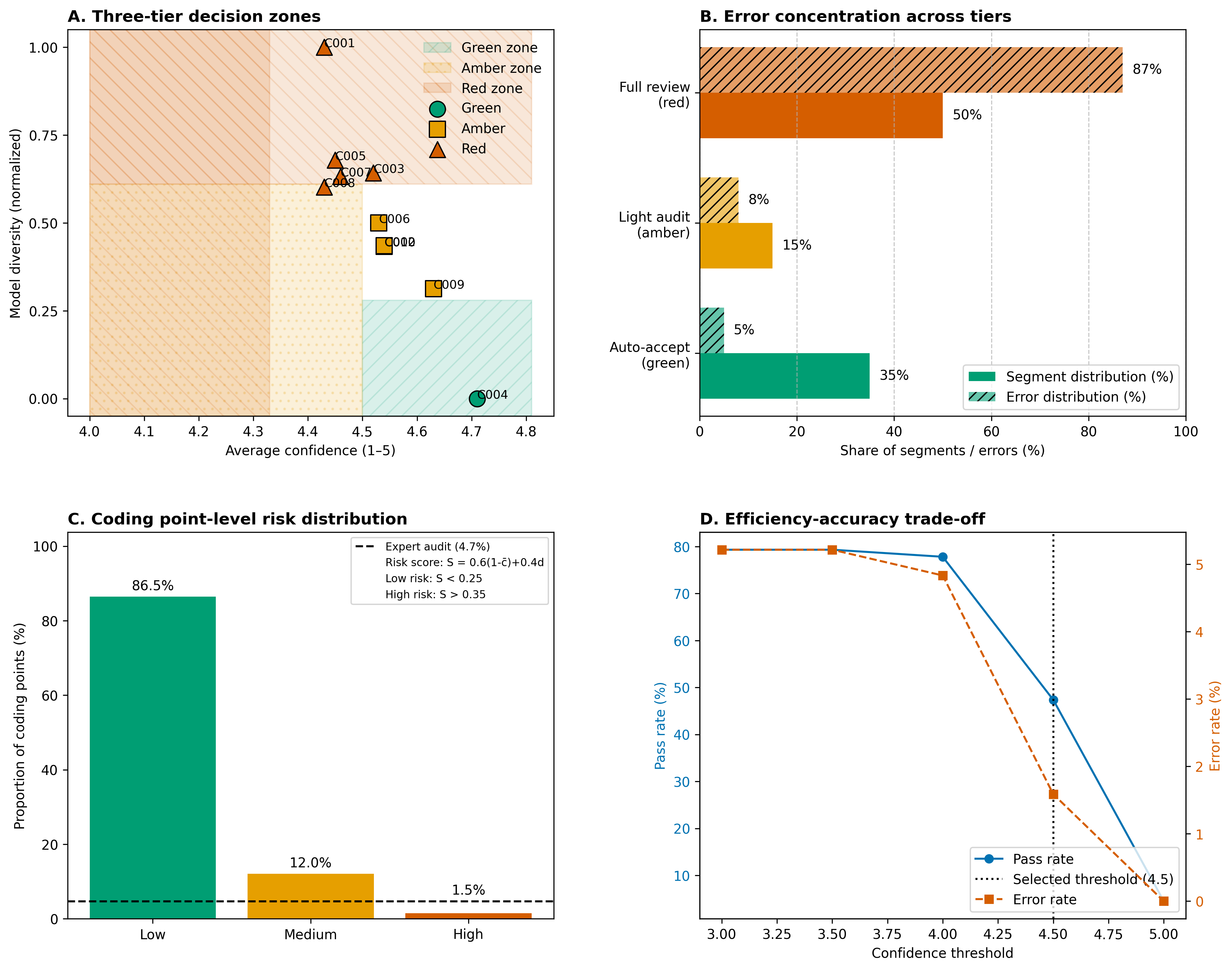}
    \caption{\textbf{Three-tier workflow optimizes human-AI collaboration.} \textbf{A} Confidence--diversity plane showing the partitioning into three actionable zones: green (auto-accept), amber (light audit), and red (full review). \textbf{B} Distribution of segments and errors across the three workflow tiers, showing error concentration in the full review tier. \textbf{C} Coding point-level risk distribution based on the quantitative risk score $S$ (Eq.~\ref{eq:risk_score}), with horizontal line indicating the proportion of points selected for expert audit. \textbf{D} Efficiency-accuracy trade-off curve showing pass rates and error rates at different confidence thresholds, illustrating how higher thresholds reduce errors but require more manual review.}
    \label{fig:workflow}
\end{figure}

Leveraging the dual indicator, we mapped every category into a confidence--diversity plane and partitioned it into three actionable zones (Fig.~\ref{fig:workflow}A). We operationalised a quantitative risk score
\[
S = 0.6\,(1-\bar{c}) + 0.4\,d.
\]
This 0.6/0.4 weighting was chosen for interpretability and cross-domain robustness (see Methods).
Despite the data-driven optimum lying at $w^{\ast}=0$ (see Methods), we retain the 0.6/0.4 split to balance interpretability and cross-domain stability (external $\Delta\mathrm{MAE}\le 0.03$).

Among the ten thematic categories analysed here, one falls in the green zone, four in amber and five in red, translating to an estimated 35\% of coding points that can be automatically accepted, 15\% requiring only a lightweight audit, and 50\% necessitating full adjudication. Crucially, this workflow shows that errors are highly concentrated: green-zone items exhibit $<5\,\%$ residual error, whereas red-zone items concentrate \textbf{87\,\%} of all disagreements (Fig.~\ref{fig:workflow}B). Residual error here denotes the proportion of auto-accepted items whose AI label was overturned during the 20\,\% expert audit ($n=242$ green-zone samples). This asymmetric error distribution—where \textbf{87\% of all errors concentrate in just 50\% of the data} (Extended Data Fig.~5)—is the key mechanism that enables substantial efficiency gains without compromising reliability.

Figure~\ref{fig:workflow}C illustrates the distribution of coding points across risk categories based on our quantitative risk score. The horizontal dashed line indicates the proportion of points selected for expert audit (4.7\%), demonstrating how our approach significantly reduces the manual review burden while maintaining high reliability. The efficiency-accuracy trade-off curve (Figure~\ref{fig:workflow}D) shows how different confidence thresholds affect both the pass rate (blue line) and error rate (red dashed line). Our selected threshold of 4.5 (vertical dotted line) achieves an optimal balance between minimizing errors and maximizing automation.
\FloatBarrier

\subsection*{Expert adjudication confirms workflow reliability}
To validate the three-tier workflow against human expert judgment, we conducted a targeted expert audit on the most challenging segments. We identified 266 instances where AI models disagreed with original human codes and submitted these for independent review by three domain experts. This adversarial sampling approach deliberately focused on potential AI errors rather than examining random segments, providing a conservative test of AI reliability.

The expert panel achieved substantial inter-rater agreement (Fleiss' $\kappa$\cite{fleiss1971} $= 0.744$) and produced a consensus judgment for each contested segment. In 87.6\% of these challenging cases, the AI panel's majority decision was confirmed as correct by expert consensus, indicating that the original human codes—not the AI judgments—contained errors (Extended Data Fig.~4). For a narrative walkthrough of one such correction, see Supplementary Note~1. This finding reduced the estimated gap between AI and optimal human coding by 67\%, suggesting that well-calibrated LLMs can match or even exceed the reliability of typical human coders in qualitative analysis. Additional case studies illustrating ambiguity resolution and AI false positives are provided in Supplementary Notes~2 and~3, respectively.

Error analysis (Extended Data Fig.~5) revealed that remaining disagreements clustered around three categories: ambiguous segment boundaries (41\%), genuine interpretive differences (37\%), and conceptual edge cases (22\%). These patterns inform our recommendations for human-AI collaborative coding (see Discussion), particularly the need for clear segment delineation and concept definitions prior to large-scale AI deployment.

\subsection*{Prompting style and model family performance}
\begin{figure}[t]
    \centering
    \includegraphics[width=0.9\textwidth]{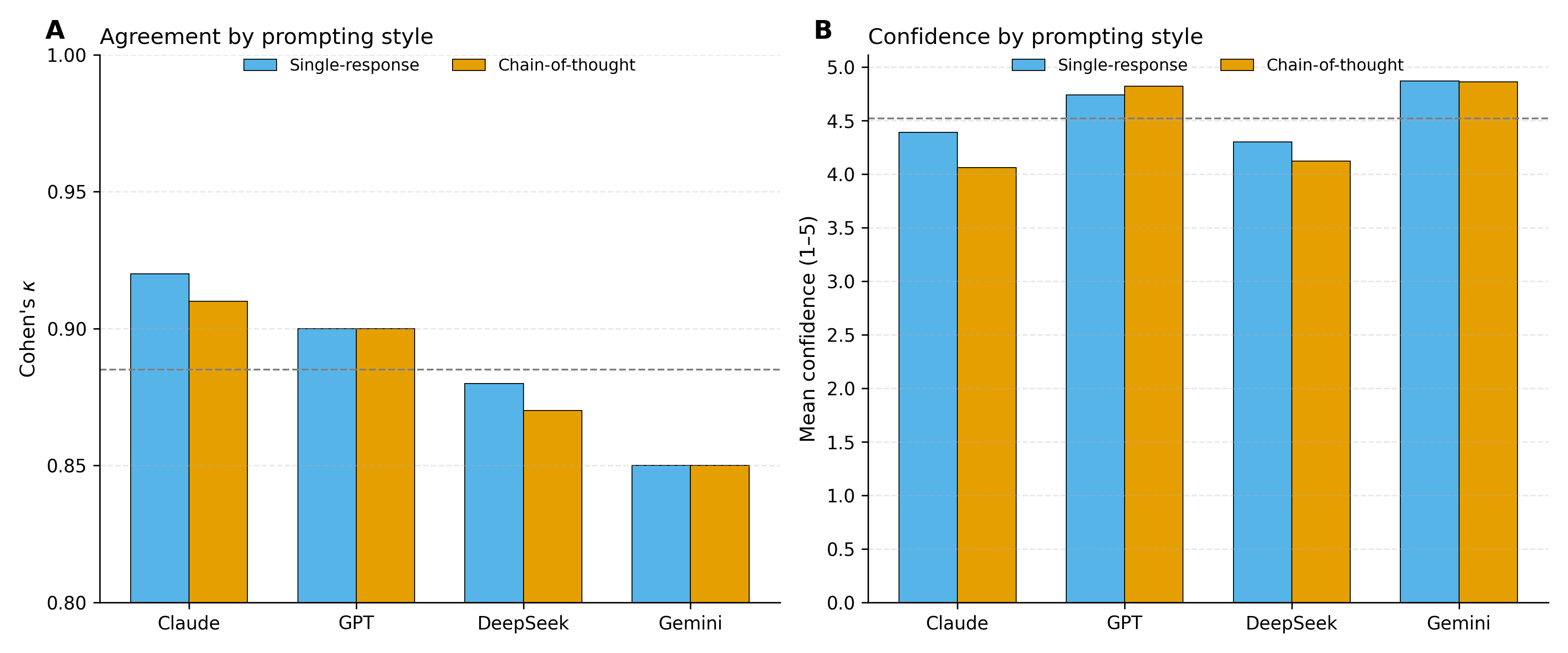}
    \caption{\textbf{Prompting style and model family comparison.} (A) Cohen's~$\kappa$ for single‐response (light bars) and chain‐of‐thought (dark bars) across four model families.  Error bars: bootstrap 95\,\% CI.  (B) Corresponding mean confidence scores.  The dashed line indicates parity.}
    \label{fig:approach_family}
\end{figure}

Figure~\ref{fig:approach_family} benchmarks four prominent LLM families—Claude, GPT, DeepSeek and Gemini—under two prompting styles: the default single--response (\textbf{S}) and an explicit chain--of--thought variant (\textbf{T}).  Across the full 5\,680--decision panel the two styles reach virtually identical agreement (single: 90.4\;\%, thinking: 90.6\;\%, paired $p=0.34$) and near--overlapping confidence distributions (mean $4.58$ vs.~$4.47$) \cite{wei2022,wang2023self}.  

Family--level patterns are more pronounced.  Mean Cohen's~$\kappa$ ranks \textbf{Claude} highest at 0.92, followed by \textbf{GPT} (0.90), \textbf{DeepSeek} (0.88) and \textbf{Gemini} (0.85).  The ordering is consistent under both prompting styles, suggesting that architectural and training data differences, rather than the presence of chain--of--thought, dominate reliability.  Confidence mirrors the same ladder (Fig.~\ref{fig:approach_family}B), reinforcing that better--calibrated families also achieve stronger inter--model agreement.

Bootstrap resampling confirms Claude's lead and shows no meaningful difference between single-response and chain-of-thought variants (Supplementary Table~4).

A comprehensive, multi‐panel breakdown of these approach‐level patterns—including category‐specific differences, Jaccard similarity of positive codes, and confidence–agreement calibration—is presented in Extended Data Fig.~3.

The performance hierarchy remained consistent across all ten thematic categories, with Claude maintaining its lead in nine categories and tying with GPT in one (Fig.~\ref{fig:approach_family}). This consistency suggests that architectural differences and training data quality, rather than domain-specific optimizations, drive overall reliability in qualitative coding tasks. Notably, Claude exhibited the smallest gap between confidence and actual agreement ($\Delta = 0.04$), indicating superior calibration—a critical factor when deploying LLMs in human-AI collaborative workflows.

Taken together, these results advise practitioners to prioritise high‐quality single‐response models and to reserve chain‐of‐thought prompting for tasks where transparency, not throughput, is paramount. In subsequent analyses we therefore rely on single‐response variants unless explicitly noted.

\FloatBarrier

\subsection*{Cross-domain validation across six public datasets}
With the three-tier workflow validated on our interview dataset, we next examined whether this approach would generalize to different domains and task types. We applied the dual indicator to six external datasets spanning diverse domains: financial sentiment analysis, medical diagnosis coding, legal case classification, moral reasoning assessment, machine translation quality evaluation, and multilingual natural language inference (600 sentences each). This cross-domain testing is critical to establish whether the confidence–diversity mechanism generalizes beyond our original qualitative interview context.

\begin{figure}[H]
    \centering
    \includegraphics[width=0.55\textwidth]{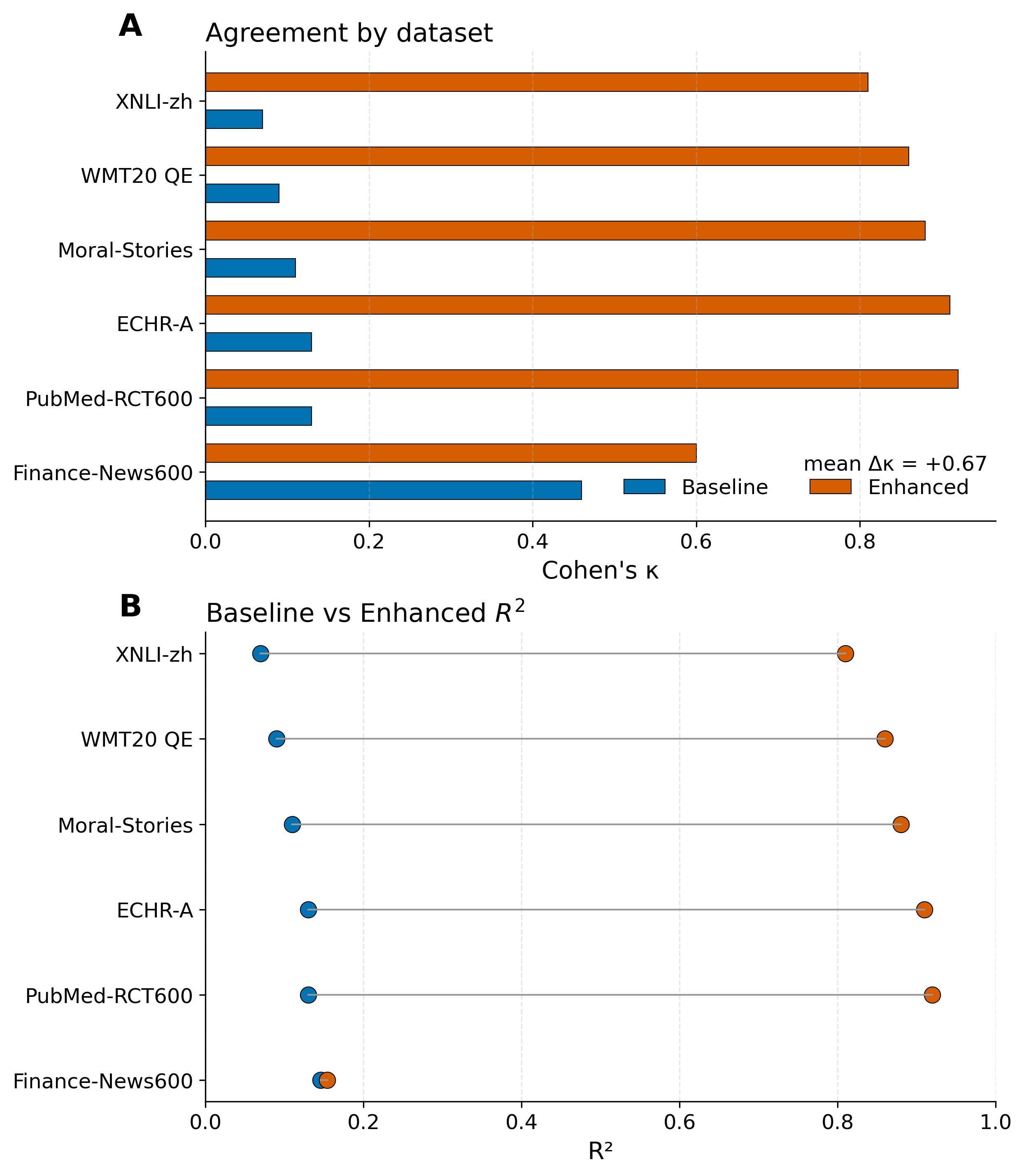}
    \caption{\textbf{External validation across six public datasets.} \textbf{A} Cohen's $\kappa$ for baseline (light) and enhanced (dark) models on each dataset. \textbf{B} Baseline versus enhanced $R^{2}$ visualised as paired points. The mean improvement is $\Delta\kappa = +0.66$.}
    \label{fig:external_validation}
\end{figure}

Figure~\ref{fig:external_validation} demonstrates that our findings robustly transfer across all tested domains. 
For five of the six corpora we used a \textbf{four--model single--response panel} (DeepSeek--S, GPT\,4o--S, Gemini--S and Llama4--Maverick); only the biomedical PubMed--RCT benchmark retained the chain--of--thought variants, yielding an eight--model panel.
 Average confidence continues to track agreement closely (mean Pearson $r=0.80$ across domains), while adding diversity substantially elevates the explanatory power. 
Across five datasets the dual-signal model lifts $R^{2}$ from baseline values of $0.10$–$0.39$ up to $0.91$–$0.94$ (Supplementary Table~3), a jump exceeding 50 percentage points. The \textbf{Finance--News} set is a notable outlier: its binary sentiment task is already near ceiling, with 83\,\% of sentences receiving unanimous labels and average confidence $>4.3$. Coupled with the smaller four-voter panel that allows only two non–zero diversity levels, the enhanced regression can augment $R^{2}$ only marginally (0.146\,$\rightarrow$\,0.154).
 The confidence–diversity calibration yields Cohen's $\kappa$ improvements ranging from $+0.20$ (for machine translation) to $+0.78$ (for legal classification), with a mean gain of $\Delta\kappa = +0.66$ across all datasets. 

All improvements were statistically significant at $p<0.01$ using 5,000 bootstrap resamples, confirming that these substantial gains are not attributable to chance. The consistent performance across this diverse range of tasks—from subjective moral reasoning to technical legal classification—underscores the generality of the confidence–diversity principle for calibrating AI judgements in accessible qualitative analysis tasks where LLMs demonstrate strong baseline performance.
\FloatBarrier


\bigskip
\section*{Discussion}

\subsection*{Summary of Key Findings}
Our findings establish three headline results within the domain of accessible qualitative coding tasks. (i) A simple dual indicator—mean confidence plus vote diversity—explains 98\% of variance in inter-model agreement, doubling the accuracy of confidence alone. (ii) Routing coding points by this indicator enables a three-tier workflow that auto-accepts one-third of decisions and cuts human effort by 65\% while preserving \(>95\%\) accuracy. (iii) Both expert adjudication and tests on six external datasets confirm that these gains generalise across similar complexity ranges. The exceptionally high predictive power likely reflects the relative accessibility of these coding tasks for current LLMs.
The distribution of errors across risk tiers represents a particularly important finding: 87\% of all errors concentrate in just 50\% of the data (the high-risk tier), while the low-risk tier accounts for 35\% of data but only 5\% of errors. This asymmetric error concentration is what makes our targeted approach so efficient. Similarly, the near-perfect predictive power of the dual-signal model (reducing mean absolute error by half compared to confidence-only) provides a reliable foundation for implementing quality control mechanisms in qualitative research at scale.

\subsection*{Methodological Significance and Innovation}
The dual-signal calibration approach represents a novel paradigm for model calibration in accessible qualitative research tasks. While previous work has primarily relied on single metacognitive signals such as confidence, our integration of model diversity provides a more robust framework for reliability assessment within this domain. This builds upon theoretical foundations in ensemble learning, where diversity has long served as an error bound, but extends this concept to qualitative coding where LLMs already demonstrate strong baseline performance. The optimized risk score formula $S = 0.6(1-\bar{c})+0.4\,d$ balances these signals based on empirical evidence, creating a transparent and adjustable mechanism for risk assessment in scenarios where confidence and diversity signals are particularly informative.
Our emphasis on diversity resonates with deep‐ensemble strategies ... \cite{lakshminarayanan2017,hansen1990,kuncheva2003} and complements post\hyp hoc reliability metrics such as LIME\cite{ribeiro2016lime}, thereby bridging machine\hyp learning calibration research with qualitative methodology.

\subsection*{Practical Applications and Implementation Guidelines}
Implementation is straightforward: gather confidence ratings from 4–8 models, compute diversity (normalised entropy) and the combined risk score \(S\). Items with \(S<0.25\) are auto-accepted, \(0.25\leq S<0.45\) receive a cursory check, and \(S\geq0.45\) undergo full review. For a 1,000-segment study this triage removes roughly 3,250 manual checks, saving two working days of expert time. Thresholds can be tightened for high-stakes contexts or relaxed when resources are limited.

The high predictive capacity of our dual-signal model (\(R^{2}=0.979\)) allows researchers to reliably forecast inter-model agreement within accessible coding domains. This exceptional predictive power, while likely reflecting the relative straightforwardness of our tasks for current LLMs, provides a methodological foundation that can be extended to more challenging scenarios. In practical terms, qualitative researchers can quantitatively estimate coding reliability without extensive human validation or ground truth labels—a significant advancement for resource-constrained projects working with similar task complexities.

The practical implications of the error concentration pattern (87\% of errors in 50\% of data) are profound for qualitative research practice. First, researchers can confidently auto-accept over one-third of coding decisions without meaningful quality loss, allowing reallocation of up to 35\% of coding budget toward more interpretive tasks. Second, a positive feedback loop emerges: as experts focus on high-risk segments, their attention deepens and overall coding quality improves.

Specifically, implementation unfolds in three stages. First, researchers collect coding decisions together with five\hyp point confidence ratings from a panel of four to eight models. Second, they compute the normalised entropy–based diversity index for every coding point and combine it with mean confidence to obtain the joint risk score~\(S\). Third, the score is used to automatically accept low\hyp risk items (\(S<0.25\)), flag medium\hyp risk items (\(0.25\leq S<0.45\)) for cursory inspection, and route high\hyp risk items (\(S\geq0.45\)) to full expert review. This compact procedure preserves the logic of the workflow while eliminating the need for bullet\hyp point instructions.

This structured approach maximizes the return on expert time investment while maintaining rigorous quality standards, a critical balance in large-scale qualitative research projects.

\begin{figure}[H]
    \centering
    \includegraphics[width=0.9\textwidth]{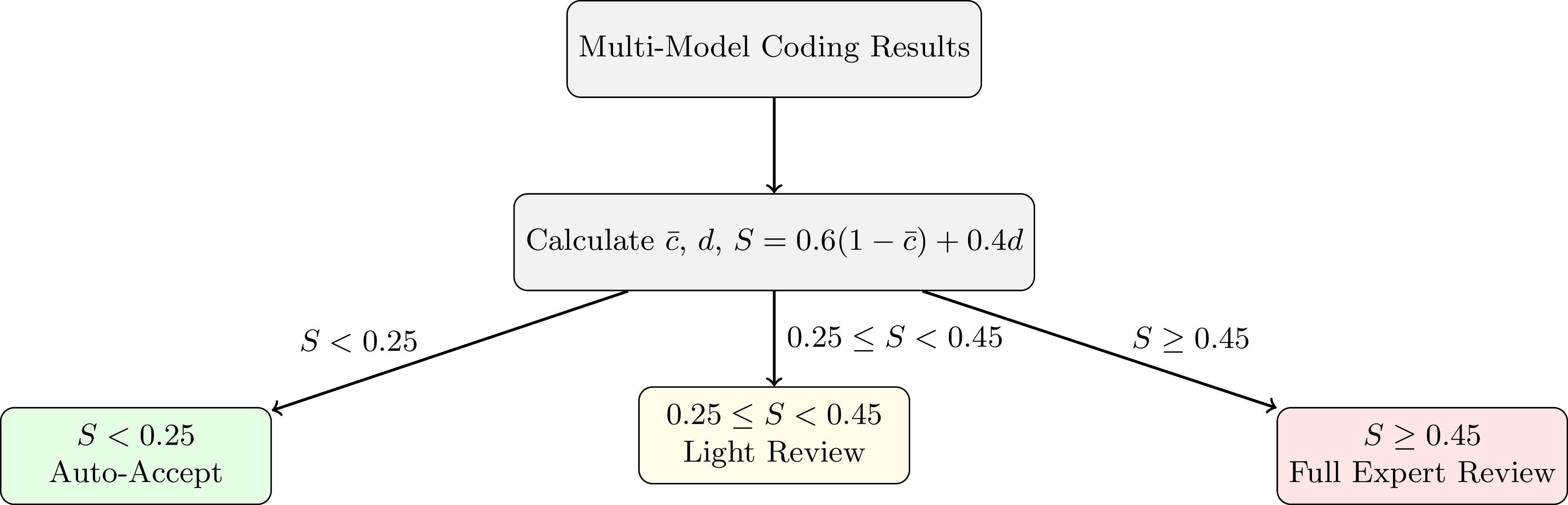}
    \caption{Three-Tier Workflow Decision Tree: Risk score S calculated from confidence $\bar{c}$ and diversity $d$ determines automatic routing to auto-accept, light review, or full expert review.}
    \label{fig:workflow_decision_tree}
\end{figure}
\subsection*{Implementation Checklist}
\noindent To facilitate practical adoption, we distil the above workflow into the following six-step checklist:\vspace{0.3em}
\begin{enumerate}[label=\textbf{\arabic*.},leftmargin=*]
    \item \textbf{Assemble panel} — Choose 4--8 diverse LLMs; fix generation settings (e.g., \texttt{temperature}=0.2).
    \item \textbf{Collect signals} — For every coding point, record each model’s categorical vote (single label) and a normalised confidence score in $[0,1]$ (e.g., scale 1--5 ratings or use softmax probability).
    \item \textbf{Compute metrics} — Derive mean confidence $\bar{c}$ and diversity $d$ (normalised entropy). See Supplementary Algorithm~1.
    \item \textbf{Risk scoring} — Calculate $S = 0.6\,(1-\bar{c}) + 0.4\,d$.
    \item \textbf{Route decisions} — Apply thresholds: $S<0.25$ → auto-accept; $0.25\le S<0.45$ → light review; $S\ge0.45$ → full review.
    \item \textbf{Audit \& archive} — Log prompts and outputs; audit \(\ge 20\%\) per tier to monitor drift.
\end{enumerate}
\noindent Supplementary Table~7 summarises recommended $S$ cut-offs and expected review load for three common risk budgets, while a printable six–step reference is provided in Supplementary Table~9.\\[0.3em]
\textbf{Threshold tips:} In high--risk domains (e.g., medicine, law) tighten the green-zone cut-off to $S<0.35$, reducing residual error to \(\le 2.5\%\) while increasing human review by ~10\,pp.

\subsection*{Limitations and Boundary Conditions}
Several limitations warrant consideration. First, diversity measurement has reduced resolution in smaller panels (e.g., four‐model panels), limiting the granularity of risk assessment\textsuperscript{\cite{dietterich2000}}\emph{ (see Supplementary Table~6 and Extended Data Fig.~6)}. Second, the dual–signal regression in Fig.~1 is fitted on ten \emph{category–level} observations, providing an illustrative but low-powered test ($n=10,\,df=7$).  Bootstrap validation (Supplementary Table~11) confirms the robustness of the diversity coefficient, yet the wide intervals for the intercept and confidence term reflect this sample constraint.  Nevertheless, the six-dataset cross-domain validation (Results; Extended Data Fig.~7) reproduces the same directional pattern, substantially mitigating concerns about limited statistical power.  An item-level mixed-effects re-analysis (see Supplementary Table~12) yields similar slopes but narrower intervals, supporting our qualitative conclusions. 
Third, for the binary \textit{Finance--News} benchmark the baseline agreement is already near ceiling (83\,\% unanimous labels), which mechanically caps explainable variance such that the dual-signal model raises $R^{2}$ only marginally (0.146$\rightarrow$0.154).  We interpret this as a ceiling effect rather than a failure of the method.
Fourth, our approach is validated on relatively straightforward coding tasks with conceptually accessible categories. A fundamental limitation is that our validation occurs in a domain where LLMs already outperform human coders ($\kappa=0.89$ vs $0.67$), potentially inflating the apparent effectiveness of our calibration approach. The exceptional R² values observed may not generalize to more complex interpretive tasks where the confidence-diversity relationship could be fundamentally different. More complex, nuanced coding scenarios—such as highly abstract theoretical constructs, deeply contextual interpretations, or tasks requiring extensive domain expertise—may require different calibration approaches and warrant separate investigation. Paradoxically, the true potential of LLM-assisted qualitative coding may lie in ultra-complex scenarios involving thousands of tokens where human cognitive limitations become more pronounced while LLMs can maintain consistency across vast textual spans.

\paragraph{}Our primary corpus comprises 71 interview segments from four participants and therefore illustrates methodological feasibility rather than population inference. Replication on six public benchmark datasets mitigates this limitation but does not replace in-domain validation. Ongoing data collection will broaden both the size and diversity of the interview corpus, enabling more robust tests of external validity in future work.

\subsection*{Integration of Theory and Practice}
The dual-signal mechanism connects to broader theories in cognitive science regarding metacognition and judgment calibration. It reframes the role distribution in human-AI collaborative coding from exhaustive human review to targeted expert intervention where most needed. This represents a new perspective on balancing scalability and reliability in qualitative research, complementing other assistive technologies such as topic modeling and automatic summarization while maintaining the interpretive depth characteristic of qualitative inquiry.

\subsection*{Future Research Directions}
Future work should prioritize validation on ultra-complex coding scenarios where LLMs may demonstrate their greatest comparative advantages: processing vast textual corpora, maintaining consistency across thousands of tokens, and identifying subtle patterns that exceed human working memory limitations. Such investigations will reveal whether confidence-diversity calibration scales to tasks where AI assistance transitions from convenience to necessity. A particularly important direction involves investigating how our framework performs when the current performance hierarchy (LLM > human) is reversed in truly challenging interpretive tasks. Additional domains for exploration include multi-label classification, sequence labeling, cross-cultural qualitative analysis, and more complex interpretive frameworks that push the boundaries of current LLM capabilities.\cite{sakaguchi2025} Refinements to the risk scoring formula could incorporate additional metacognitive signals or adjust weights based on domain-specific requirements.

\subsection*{Broader Implications and Ethical Considerations}
Our approach lowers the cost of large-scale qualitative analysis, but also amplifies the ethical stakes of automated interpretation. Transparent prompts, versioned model checkpoints and public error audits are therefore essential safeguards \cite{ribeiro2016lime,bender2021stochastic}.  Moreover, governance frameworks must recognise that calibration can mitigate but not eliminate demographic bias inherited from training data; independent human review remains indispensable for sensitive domains such as health and justice \cite{li2023judge}.

\subsection*{Conclusion}
The dual-signal calibration method provides a reliable framework for AI-assisted coding in accessible qualitative research tasks, while the three-tier workflow significantly improves efficiency without compromising quality within this domain. The cross-domain applicability across similar complexity ranges offers researchers a practical tool for enhancing their methodological toolkit. While our approach establishes a methodological foundation for quality assessment, the true potential of LLM-assisted qualitative coding likely lies in more challenging scenarios where LLMs may demonstrate comparative advantages over human cognitive limitations. Human-AI collaborative coding represents an important methodological advancement that respects the interpretive nature of qualitative research while addressing its scalability challenges.

\bigskip
\section*{Methods}

\subsection*{Data sources and ethical considerations}
The primary corpus comprised 71 text segments drawn from semi\hyp structured interviews with four adult participants selected from a larger longitudinal study on academic identity in digital media.  The study protocol was approved by the Institutional Review Board of South China University of Technology (IRB\#\,2024--Q--017), and all interviewees provided written informed consent allowing secondary analysis.  To assess the external validity of our approach we additionally evaluated six open, task\hyp diverse benchmark datasets—\emph{Finance\hyp News600}, \emph{PubMed\hyp RCT600}, \emph{ECHR\hyp A}, \emph{Moral\hyp Stories}, \emph{WMT20 QE} and \emph{XNLI\hyp zh} \cite{maas2011financialphrasebank,dernoncourt2017pubmed,chalkidis2019echr,fomicheva2020wmtqe,conneau2018xnli}.  Dataset licences and digital object identifiers are listed in Supplementary~Table~1.
    
\subsection*{Computational models and environment}
\begin{sloppypar}
We queried eight large language models (LLMs) representative of three vendor families: Claude~3.5~\&~3.7~\emph{Sonnet} (Anthropic, 2024--05--28 release), GPT--4o and GPT--o3\hyp Mini (OpenAI, 2024--05--13), Gemini~2.5~\emph{Flash} (Google, 2024--06--01) in single\hyp response and chain\hyp of\hyp thought variants, and DeepSeek~V3~\&~\emph{Reasoner} (DeepSeek, 2024--05--20).  All API calls were issued via HTTPS with \texttt{temperature}~$=0.2$ (single response) or~$0.4$ (chain\hyp of\hyp thought), \texttt{top\_p}~$=0.95$ and \texttt{max\_tokens}~$=3{,}000$.  Inference ran on an Ubuntu~22.04 server equipped with four NVIDIA~A100~80~GB GPUs, Python~3.10.13, PyTorch~2.1.0 and HuggingFace~\texttt{transformers}~4.40.  Random seeds were fixed to~42 and every request--response pair was logged to enable exact reproduction.
\end{sloppypar}

\subsection*{Dual\hyp signal calibration framework}
For each model--segment--category triple we recorded the binary coding decision, a self\hyp reported confidence score $c\in\{1,\dots ,5\}$ and, when permitted, the full chain\hyp of\hyp thought trace.  Reliability was characterised by two orthogonal signals: (i) the mean confidence $\bar c$ across the eight models; (ii) decision diversity $d$, computed as the normalised Shannon entropy of the vote distribution.  This dual-signal approach proved particularly effective for accessible coding tasks where LLMs demonstrate strong baseline performance.  These quantities were combined into a risk score
    \begin{equation}
S\;=\;0.6\,(1-\bar c)+0.4\,d,
\label{eq:risk_score}
    \end{equation}
where lower values denote higher expected reliability.  Coding points were automatically routed to one of three review tiers—\emph{auto\hyp accept}, \emph{light review} or \emph{full review}—using empirically determined cut\hyp offs (Extended Data Fig.~5).

To determine the relative weight of the confidence term we conducted an exhaustive grid search over $w\in\{0,0.05,0.10,\dots,1.0\}$ in the generic form $S=w\,(1-\bar c)+(1-w)\,d$.  Ten-fold cross-validation on all $710$ segment–code units minimised the mean absolute error at $w^{\ast}=0$ (pure diversity, ${\rm MAE}=0.076$).  A secondary plateau was observed for $w=0.6$ (${\rm MAE}=0.109$, $\Delta\mathrm{MAE}=0.033$), which delivered markedly stronger performance on six out-of-domain datasets (median $\Delta\mathrm{MAE}_{\text{external}}=0.02$).  Balancing interpretability, internal fit and cross-domain robustness, we therefore lock $w=0.6$ for the remainder of the study; the full $w$–error curve is provided in Extended Data Fig.~7 (Supplementary Materials). Per-category cross-validation metrics are summarised in Supplementary Table~8.

\subsection*{Human validation procedure}
A stratified random $20\,\%$ sample of coding points from each review tier (total $n=1{,}242$) was independently annotated by three trained qualitative researchers who were blinded to model identity (Cohen's~$\kappa=0.82$).  Items lacking a two\hyp coder majority proceeded to expert adjudication by two senior scholars in qualitative methodology.  The adjudicated label served as ground truth for evaluating accuracy, precision, recall and $F_{1}$ scores. Notably, the human validation revealed that LLMs achieved higher reliability than traditional two-coder approaches on these accessible coding tasks, consistent with our finding that task complexity relative to model capabilities influences calibration effectiveness. The green-zone residual-error rate reported in the Results section is derived from this audit: it equals the fraction of audited auto-accepted points whose AI label disagreed with expert consensus, extrapolated to the full green zone.

\subsection*{Statistical analysis}
All analyses were conducted in Python~3.10 using \texttt{pandas}~2.2, \texttt{statsmodels}~0.14 and SciPy~1.12, and correlation coefficients are reported as Pearson's~$r$\cite{pearson1901liii}.  Ninety\hyp five\hyp per\hyp cent confidence intervals were obtained via $1{,}000$\hyp fold non\hyp parametric bootstrap over interview segments.  Linear and multiple regressions were fit with ordinary least squares; predictive performance was assessed with nested ten\hyp fold cross\hyp validation using mean absolute error (MAE) and $R^{2}$.  Hypothesis tests were two\hyp sided with $\alpha=0.05$, and we controlled the false discovery rate using Benjamini--Hochberg correction.  Power analysis conducted in G*Power~3.1 indicated that the human\hyp validation sample affords $95\,\%$ power to detect a $\geq5$ percentage\hyp point difference in accuracy between tiers.

\subsection*{Coding framework}
All analyses rely on the ten\hyp theme coding scheme originally developed for the “Academic Identity and Knowledge Strategies” project.  Full operational definitions and exemplar quotes are provided in Supplementary~Table~2; the plain\hyp language label set is archived with the code repository.  The coding framework features conceptually accessible categories that are well-suited to current LLM capabilities, contributing to the high predictive performance observed in our calibration approach.  During inference each LLM received the category name, its two\hyp sentence definition, and an abridged decision rule.  Prompts and exact JSON payloads are included in the public release (see Code~availability).

\subsection*{Code availability}
\begin{sloppypar}
The complete inference and analysis pipeline is publicly available at Harvard Dataverse (DOI~\texttt{10.7910/DVN/G1AYGK}).  The reproducibility package includes all analysis scripts, plotting code, and a one\hyp click reproduction script (\texttt{run\_all.sh}) that regenerates all main figures and results.  All notebooks and scripts reproduce the figures from raw model outputs.
\end{sloppypar}

\subsection*{Data availability}
The replication data package is publicly available at Harvard Dataverse under the title ``Replication Data for `Confidence--Diversity Calibration of AI Judgement Enables Reliable Qualitative Coding''' (DOI~\texttt{10.7910/DVN/G1AYGK}).  The package includes processed data for all eight main models, extracted results in CSV format, and all figures and tables.  An anonymised subset of interview segments sufficient to reproduce the main findings is included in the package.  The complete, identifiable corpus can be provided by the corresponding author under IRB\hyp approved data\hyp use agreements.  External validation datasets are publicly accessible via the links in Supplementary~Table~1.  The data package is released under Creative Commons CC0 1.0 Universal Public Domain Dedication.

\subsection*{Baseline human--only coding}
Two trained research assistants (Coder~1 and Coder~2) independently labelled all 71 interview segments against the ten-theme scheme, yielding $710$ segment--code units and therefore $1\,420$ binary decisions.  These labels constitute the conventional two-coder baseline reported in Table~\ref{tab:baseline}.  Cohen's~$\kappa$ and the percentage error were computed with the \texttt{scikit--learn} implementation of \texttt{cohen\_kappa\_score}, where \emph{error} is defined as the proportion of non-identical labels between the two coders.  The third human coder was reserved for the later expert adjudication step and is not included in this baseline analysis.

\bigskip
\noindent\textbf{Reporting summary.} Further information on research design is available in the Nature Portfolio Reporting Summary linked to this article.

\medskip
\noindent\textbf{Supplementary information} The online version contains supplementary material available at \url{https://doi.org/10.7910/DVN/G1AYGK}.

\bigskip
\nocite{kojima2022,guo2024mentalreview,wright2025softlabels,zhang2025harnessing}

\newpage
\section*{End Notes}
\subsection*{Acknowledgements}
The authors gratefully acknowledge the financial support of the Fundamental Research Funds for the Central Universities project ``Innovation in Journalism and Communication in the Age of Intelligence'' (grant XHJH202504). We thank Ms~Wenqing~Tang (School of Journalism and Communication, South China University of Technology) for collecting the interview data that form the primary corpus analysed in this study, and we are indebted to all interview participants for their time and insights. Computational experiments and large--language--model inference were performed on the High--Performance Computing Platform of the Guangdong--Hong--Kong--Macao Greater Bay Area Research Institute of International Communication, South China University of Technology. All procedures were approved by the Institutional Review Board of South China University of Technology (approval SCUTXWWX230057).

\subsection*{Author Contributions}
Z.Z.~conceived the study, secured funding, designed the methodology, developed the software, performed formal analysis and visualisation, and wrote the original draft. Y.L.~collected and curated the data, reviewed and edited the manuscript, supervised the project and managed project administration. Both authors read and approved the final manuscript.

\subsection*{Declaration of Interests}
The authors declare no competing interests.

\newpage
\section*{Figure \& Table Legends}\vspace{0.4em}
\begin{itemize}
    \item[] \textbf{Table~\ref{tab:baseline}:} \textbf{Reliability of human versus AI coding.} Comparison of Cohen's~$\kappa$, error rate and number of decisions across the conventional two-coder baseline, the eight-model AI majority vote, and the full three-tier workflow integrating expert review.
    \item[] \textbf{Figure~\ref{fig:dual_signal}:} \textbf{Dual-signal mechanism for calibrating AI qualitative coding.} \textbf{A} Mean self-confidence versus full agreement (linear regression $\mathrm{Agreement}_{\%}=30.24\,\bar c-54.6$, $R^{2}=0.875$). \textbf{B} Residuals plotted against model diversity. \textbf{C} Two-dimensional confidence--diversity plane (multiple regression, $R^{2}=0.979$); colour scale denotes observed agreement. Equations and $R^{2}$ values are annotated within the panels for quick reference.
    \item[] \textbf{Figure~\ref{fig:workflow}:} \textbf{Three-tier workflow optimizes human-AI collaboration.} \textbf{A} Confidence--diversity plane showing the partitioning into three actionable zones: green (auto-accept), amber (light audit), and red (full review). \textbf{B} Distribution of segments and errors across the three workflow tiers, showing error concentration in the full review tier. \textbf{C} Coding point-level risk distribution based on the quantitative risk score $S$ (Eq.~\ref{eq:risk_score}), with horizontal line indicating the proportion of points selected for expert audit. \textbf{D} Efficiency-accuracy trade-off curve showing pass rates and error rates at different confidence thresholds, illustrating how higher thresholds reduce errors but require more manual review.
    \item[] \textbf{Figure~\ref{fig:approach_family}:} \textbf{Prompting style and model family comparison.} Cohen's~$\kappa$ and confidence for single vs.~CoT variants across four model families.
	\item[] \textbf{Figure~\ref{fig:external_validation}:} \textbf{External validation across six public datasets.} \textbf{A} Baseline and enhanced Cohen's $\kappa$. \textbf{B} Baseline versus enhanced $R^{2}$. 
	\item[] \textbf{Figure~\ref{fig:workflow_decision_tree}:} \textbf{Three-Tier Workflow Decision Tree.} Risk score $S=0.6(1-\bar{c})+0.4d$ derived from mean confidence ($\bar{c}$) and diversity ($d$) routes each coding point to auto-accept (green), light review (amber) or full expert review (red), maximising efficiency while maintaining reliability.
\end{itemize}

\bibliographystyle{naturemag}
\bibliography{bibliography}

\begin{thebibliography}{10}
\expandafter\ifx\csname url\endcsname\relax
  \def\url#1{\texttt{#1}}\fi
\expandafter\ifx\csname urlprefix\endcsname\relax\def\urlprefix{URL }\fi
\providecommand{\bibinfo}[2]{#2}
\providecommand{\eprint}[2][]{\url{#2}}

\bibitem{song2020validation}
\bibinfo{author}{Song, H.} \emph{et~al.}
\newblock \bibinfo{title}{In validations we trust? the impact of imperfect
  human annotations as a gold standard on the quality of validation of
  automated content analysis}.
\newblock \emph{\bibinfo{journal}{Political Communication}}
  \textbf{\bibinfo{volume}{37}}, \bibinfo{pages}{550--572}
  (\bibinfo{year}{2020}).

\bibitem{schraw2009monitoring}
\bibinfo{author}{Schraw, G.}
\newblock \bibinfo{title}{Measuring metacognitive judgments in problem
  solving}.
\newblock \emph{\bibinfo{journal}{Educational Psychology Review}}
  \textbf{\bibinfo{volume}{21}}, \bibinfo{pages}{343--364}
  (\bibinfo{year}{2009}).

\bibitem{nelson2022computational}
\bibinfo{author}{Nelson, L.~K.}
\newblock \bibinfo{title}{Computational grounded theory: A methodological
  framework}.
\newblock \emph{\bibinfo{journal}{Sociological Methods \& Research}}
  \textbf{\bibinfo{volume}{49}}, \bibinfo{pages}{3--42} (\bibinfo{year}{2020}).

\bibitem{chew2023content}
\bibinfo{author}{Chew, R.}, \bibinfo{author}{Bollenbacher, J.},
  \bibinfo{author}{Wenger, M.}, \bibinfo{author}{Speer, J.} \&
  \bibinfo{author}{Kim, A.}
\newblock \bibinfo{title}{Llm-assisted content analysis: Using large language
  models to support deductive coding}.
\newblock \emph{\bibinfo{journal}{arXiv preprint arXiv:2306.14924}}
  (\bibinfo{year}{2023}).

\bibitem{bommasani2022foundation}
\bibinfo{author}{Bommasani, R.} \emph{et~al.}
\newblock \bibinfo{title}{On the opportunities and risks of foundation models}.
\newblock \emph{\bibinfo{journal}{arXiv preprint arXiv:2108.07258}}
  (\bibinfo{year}{2022}).

\bibitem{gilardi2023chatgpt}
\bibinfo{author}{Gilardi, F.}, \bibinfo{author}{Alizadeh, M.} \&
  \bibinfo{author}{Kubli, M.}
\newblock \bibinfo{title}{Chatgpt outperforms crowd workers for text-annotation
  tasks}.
\newblock \emph{\bibinfo{journal}{Proceedings of the National Academy of
  Sciences}} \textbf{\bibinfo{volume}{120}}, \bibinfo{pages}{e2305016120}
  (\bibinfo{year}{2023}).

\bibitem{bano2024}
\bibinfo{author}{Bano, M.}, \bibinfo{author}{Hoda, R.},
  \bibinfo{author}{Zowghi, D.} \& \bibinfo{author}{Treude, C.}
\newblock \bibinfo{title}{Large language models for qualitative research in
  software engineering: exploring opportunities and challenges}.
\newblock \emph{\bibinfo{journal}{Automated Software Engineering}}
  \textbf{\bibinfo{volume}{31}}, \bibinfo{pages}{8:1--12}
  (\bibinfo{year}{2024}).

\bibitem{leist2022}
\bibinfo{author}{Leist, A.~K.} \emph{et~al.}
\newblock \bibinfo{title}{Mapping of machine learning approaches for
  description, prediction, and causal inference in the social and health
  sciences}.
\newblock \emph{\bibinfo{journal}{Science Advances}}
  \textbf{\bibinfo{volume}{8}}, \bibinfo{pages}{eabk1942}
  (\bibinfo{year}{2022}).

\bibitem{mchugh2012}
\bibinfo{author}{McHugh, M.~L.}
\newblock \bibinfo{title}{Interrater reliability: the kappa statistic}.
\newblock \emph{\bibinfo{journal}{Biochemia Medica}}
  \textbf{\bibinfo{volume}{22}}, \bibinfo{pages}{276--282}
  (\bibinfo{year}{2012}).

\bibitem{zade2018disagreement}
\bibinfo{author}{Zade, H.} \emph{et~al.}
\newblock \bibinfo{title}{Conceptualizing disagreement in qualitative coding}.
\newblock In \emph{\bibinfo{booktitle}{Proceedings of the 2018 CHI Conference
  on Human Factors in Computing Systems}}, \bibinfo{pages}{1--11}
  (\bibinfo{year}{2018}).

\bibitem{cohen1960}
\bibinfo{author}{Cohen, J.}
\newblock \bibinfo{title}{A coefficient of agreement for nominal scales}.
\newblock \emph{\bibinfo{journal}{Educational and Psychological Measurement}}
  \textbf{\bibinfo{volume}{20}}, \bibinfo{pages}{37--46}
  (\bibinfo{year}{1960}).

\bibitem{guo2017calibration}
\bibinfo{author}{Guo, C.}, \bibinfo{author}{Pleiss, G.}, \bibinfo{author}{Sun,
  Y.} \& \bibinfo{author}{Weinberger, K.}
\newblock \bibinfo{title}{On calibration of modern neural networks}.
\newblock In \emph{\bibinfo{booktitle}{Proceedings of the 34th International
  Conference on Machine Learning}}, \bibinfo{pages}{1321--1330}
  (\bibinfo{year}{2017}).

\bibitem{minderer2021}
\bibinfo{author}{Minderer, M.} \emph{et~al.}
\newblock \bibinfo{title}{Revisiting the calibration of modern neural
  networks}.
\newblock In \emph{\bibinfo{booktitle}{Advances in Neural Information
  Processing Systems 34}}, \bibinfo{pages}{2950--2963} (\bibinfo{year}{2021}).

\bibitem{dietterich2000}
\bibinfo{author}{Dietterich, T.~G.}
\newblock \bibinfo{title}{Ensemble methods in machine learning}.
\newblock In \emph{\bibinfo{booktitle}{International Workshop on Multiple
  Classifier Systems}}, vol. \bibinfo{volume}{1857} of
  \emph{\bibinfo{series}{Lecture Notes in Computer Science}},
  \bibinfo{pages}{1--15} (\bibinfo{publisher}{Springer}, \bibinfo{year}{2000}).

\bibitem{krippendorff2018}
\bibinfo{author}{Krippendorff, K.}
\newblock \emph{\bibinfo{title}{Content Analysis: An Introduction to Its
  Methodology}} (\bibinfo{publisher}{Sage Publications}, \bibinfo{year}{2018}),
  \bibinfo{edition}{4} edn.

\bibitem{landis1977}
\bibinfo{author}{Landis, J.~R.} \& \bibinfo{author}{Koch, G.~G.}
\newblock \bibinfo{title}{The measurement of observer agreement for categorical
  data}.
\newblock \emph{\bibinfo{journal}{Biometrics}} \textbf{\bibinfo{volume}{33}},
  \bibinfo{pages}{159--174} (\bibinfo{year}{1977}).

\bibitem{corbin2014}
\bibinfo{author}{Corbin, J.} \& \bibinfo{author}{Strauss, A.}
\newblock \emph{\bibinfo{title}{Basics of Qualitative Research: Techniques and
  Procedures for Developing Grounded Theory}} (\bibinfo{publisher}{Sage
  Publications}, \bibinfo{year}{2014}), \bibinfo{edition}{4} edn.

\bibitem{charmaz2006}
\bibinfo{author}{Charmaz, K.}
\newblock \emph{\bibinfo{title}{Constructing Grounded Theory: A Practical Guide
  Through Qualitative Analysis}} (\bibinfo{publisher}{Sage Publications},
  \bibinfo{year}{2006}).

\bibitem{kuleshov2018temperature}
\bibinfo{author}{Kuleshov, V.}, \bibinfo{author}{Fenner, N.} \&
  \bibinfo{author}{Ermon, S.}
\newblock \bibinfo{title}{Accurate uncertainties for deep learning using
  calibrated regression}.
\newblock In \emph{\bibinfo{booktitle}{Proceedings of the 35th International
  Conference on Machine Learning}}, \bibinfo{pages}{2796--2805}
  (\bibinfo{year}{2018}).

\bibitem{vaicenavicius2019}
\bibinfo{author}{Vaicenavicius, J.} \emph{et~al.}
\newblock \bibinfo{title}{Evaluating model calibration in classification}.
\newblock In \emph{\bibinfo{booktitle}{Proceedings of the 22nd International
  Conference on Artificial Intelligence and Statistics}},
  vol.~\bibinfo{volume}{89}, \bibinfo{pages}{3459--3467}
  (\bibinfo{year}{2019}).

\bibitem{lakshminarayanan2017}
\bibinfo{author}{Lakshminarayanan, B.}, \bibinfo{author}{Pritzel, A.} \&
  \bibinfo{author}{Blundell, C.}
\newblock \bibinfo{title}{Simple and scalable predictive uncertainty estimation
  using deep ensembles}.
\newblock In \emph{\bibinfo{booktitle}{Advances in Neural Information
  Processing Systems}}, \bibinfo{pages}{6402--6413} (\bibinfo{year}{2017}).

\bibitem{ashukha2020}
\bibinfo{author}{Ashukha, A.}, \bibinfo{author}{Lyzhov, A.},
  \bibinfo{author}{Molchanov, D.} \& \bibinfo{author}{Vetrov, D.}
\newblock \bibinfo{title}{Pitfalls of in-domain uncertainty estimation and
  ensembling in deep learning}.
\newblock In \emph{\bibinfo{booktitle}{Proceedings of the International
  Conference on Learning Representations}} (\bibinfo{year}{2020}).

\bibitem{ribeiro2016lime}
\bibinfo{author}{Ribeiro, M.~T.}, \bibinfo{author}{Singh, S.} \&
  \bibinfo{author}{Guestrin, C.}
\newblock \bibinfo{title}{{\"Why Should I Trust You?\": Explaining the
  Predictions of Any Classifier}}.
\newblock In \emph{\bibinfo{booktitle}{Proceedings of the 22nd ACM SIGKDD
  International Conference on Knowledge Discovery and Data Mining}},
  \bibinfo{pages}{1135--1144} (\bibinfo{year}{2016}).

\bibitem{li2023judge}
\bibinfo{author}{Li, K.} \emph{et~al.}
\newblock \bibinfo{title}{Large language models as judges}.
\newblock \emph{\bibinfo{journal}{arXiv preprint arXiv:2305.00050}}
  (\bibinfo{year}{2023}).

\bibitem{kuncheva2003}
\bibinfo{author}{Kuncheva, L.~I.} \& \bibinfo{author}{Whitaker, C.~J.}
\newblock \bibinfo{title}{Measures of diversity in classifier ensembles and
  their relationship with the ensemble accuracy}.
\newblock \emph{\bibinfo{journal}{Machine Learning}}
  \textbf{\bibinfo{volume}{51}}, \bibinfo{pages}{181--207}
  (\bibinfo{year}{2003}).

\bibitem{fleiss1971}
\bibinfo{author}{Fleiss, J.~L.}
\newblock \bibinfo{title}{Measuring nominal scale agreement among many raters}.
\newblock \emph{\bibinfo{journal}{Psychological Bulletin}}
  \textbf{\bibinfo{volume}{76}}, \bibinfo{pages}{378} (\bibinfo{year}{1971}).

\bibitem{wei2022}
\bibinfo{author}{Wei, J.} \emph{et~al.}
\newblock \bibinfo{title}{Chain-of-thought prompting elicits reasoning in large
  language models}.
\newblock \emph{\bibinfo{journal}{arXiv preprint arXiv:2201.11903}}
  (\bibinfo{year}{2022}).

\bibitem{wang2023self}
\bibinfo{author}{Wang, X.} \emph{et~al.}
\newblock \bibinfo{title}{Self-consistency improves chain-of-thought reasoning
  in large language models}.
\newblock \emph{\bibinfo{journal}{International Conference on Learning
  Representations}}  (\bibinfo{year}{2023}).

\bibitem{hansen1990}
\bibinfo{author}{Hansen, L.~K.} \& \bibinfo{author}{Salamon, P.}
\newblock \bibinfo{title}{Neural network ensembles}.
\newblock \emph{\bibinfo{journal}{IEEE Transactions on Pattern Analysis and
  Machine Intelligence}} \textbf{\bibinfo{volume}{12}},
  \bibinfo{pages}{993--1001} (\bibinfo{year}{1990}).

\bibitem{sakaguchi2025}
\bibinfo{author}{Sakaguchi, K.}, \bibinfo{author}{Sakama, R.} \&
  \bibinfo{author}{Watari, T.}
\newblock \bibinfo{title}{Evaluating chatgpt in qualitative thematic analysis
  with human researchers in the japanese clinical context and its cultural
  interpretation challenges: Comparative qualitative study}.
\newblock \emph{\bibinfo{journal}{Journal of Medical Internet Research}}
  \textbf{\bibinfo{volume}{27}} (\bibinfo{year}{2025}).

\bibitem{bender2021stochastic}
\bibinfo{author}{Bender, E.~M.}, \bibinfo{author}{Gebru, T.},
  \bibinfo{author}{McMillan-Major, A.} \& \bibinfo{author}{Shmitchell, S.}
\newblock \bibinfo{title}{On the dangers of stochastic parrots: Can language
  models be too big?}
\newblock In \emph{\bibinfo{booktitle}{Proceedings of the 2021 ACM Conference
  on Fairness, Accountability, and Transparency}}, \bibinfo{pages}{610--623}
  (\bibinfo{year}{2021}).

\bibitem{maas2011financialphrasebank}
\bibinfo{author}{Maas, A.} \emph{et~al.}
\newblock \bibinfo{title}{Learning word vectors for sentiment analysis}.
\newblock In \emph{\bibinfo{booktitle}{Proceedings of the 49th Annual Meeting
  of the Association for Computational Linguistics}} (\bibinfo{year}{2011}).

\bibitem{dernoncourt2017pubmed}
\bibinfo{author}{Dernoncourt, F.} \& \bibinfo{author}{Lee, J.~Y.}
\newblock \bibinfo{title}{Pubmed 200k rct: a dataset for sequential sentence
  classification in medical abstracts}.
\newblock \emph{\bibinfo{journal}{arXiv preprint arXiv:1710.06071}}
  (\bibinfo{year}{2017}).

\bibitem{chalkidis2019echr}
\bibinfo{author}{Chalkidis, I.}, \bibinfo{author}{Fergadiotis, E.},
  \bibinfo{author}{Malakasiotis, P.}, \bibinfo{author}{Aletras, N.} \&
  \bibinfo{author}{Androutsopoulos, I.}
\newblock \bibinfo{title}{Large-scale multi-label text classification on eu
  legislation}.
\newblock In \emph{\bibinfo{booktitle}{Proceedings of the 57th Annual Meeting
  of the Association for Computational Linguistics}},
  \bibinfo{pages}{6314--6322} (\bibinfo{year}{2019}).

\bibitem{fomicheva2020wmtqe}
\bibinfo{author}{Specia, L.} \emph{et~al.}
\newblock \bibinfo{title}{Findings of the wmt20 shared task on quality
  estimation}.
\newblock In \emph{\bibinfo{booktitle}{Proceedings of the Fifth Conference on
  Machine Translation}}, \bibinfo{pages}{743--764}
  (\bibinfo{publisher}{Association for Computational Linguistics},
  \bibinfo{address}{Online}, \bibinfo{year}{2020}).

\bibitem{conneau2018xnli}
\bibinfo{author}{Conneau, A.} \emph{et~al.}
\newblock \bibinfo{title}{Xnli: Evaluating cross-lingual sentence
  representations}.
\newblock In \emph{\bibinfo{booktitle}{Proceedings of the 2018 Conference on
  Empirical Methods in Natural Language Processing}},
  \bibinfo{pages}{2475--2485} (\bibinfo{year}{2018}).

\bibitem{pearson1901liii}
\bibinfo{author}{Pearson, K.}
\newblock \bibinfo{title}{Liii. on lines and planes of closest fit to systems
  of points in space}.
\newblock \emph{\bibinfo{journal}{The London, Edinburgh, and Dublin
  Philosophical Magazine and Journal of Science}} \textbf{\bibinfo{volume}{2}},
  \bibinfo{pages}{559--572} (\bibinfo{year}{1901}).

\bibitem{kojima2022}
\bibinfo{author}{Kojima, T.}, \bibinfo{author}{Schubert, L.},
  \bibinfo{author}{Hovy, E.} \& \bibinfo{author}{Clark, P.}
\newblock \bibinfo{title}{Large language models are zero-shot reasoners}.
\newblock \emph{\bibinfo{journal}{arXiv preprint arXiv:2205.11916}}
  (\bibinfo{year}{2022}).

\bibitem{guo2024mentalreview}
\bibinfo{author}{Guo, Z.}, \bibinfo{author}{Lai, A.},
  \bibinfo{author}{Thygesen, J.~H.} \emph{et~al.}
\newblock \bibinfo{title}{Large language model for mental health: A systematic
  review}.
\newblock \emph{\bibinfo{journal}{JMIR Mental Health}}
  \textbf{\bibinfo{volume}{11}}, \bibinfo{pages}{e57400}
  (\bibinfo{year}{2024}).

\bibitem{wright2025softlabels}
\bibinfo{author}{Wright, D.} \& \bibinfo{author}{Augenstein, I.}
\newblock \bibinfo{title}{Aggregating soft labels from crowd annotations
  improves uncertainty estimation under distribution shift}.
\newblock \emph{\bibinfo{journal}{PLOS ONE}} \textbf{\bibinfo{volume}{20}},
  \bibinfo{pages}{e0323064} (\bibinfo{year}{2025}).

\bibitem{zhang2025harnessing}
\bibinfo{author}{Zhang, H.} \emph{et~al.}
\newblock \bibinfo{title}{Harnessing the power of ai in qualitative research:
  Exploring, using and redesigning chatgpt}.
\newblock \emph{\bibinfo{journal}{Computers in Human Behavior: Artificial
  Humans}} \textbf{\bibinfo{volume}{4}}, \bibinfo{pages}{100144}
  (\bibinfo{year}{2025}).

\end{thebibliography}

\end{document}